**Title:** Application of Artificial Intelligence in the Management of Climate Change-Related Health Issues in Africa

**Authors:**
**Peter Kokol [1,2*], peter.kokol@um.si**

**Martha Kachweka3, mkachwek@andrew.cmu.edu**

**Helena Blažun Vošner [1,4], Helena.blazun@zd-mb.si**

**Samuel Chol Buol[3], sbuol@andrew.cmu.edu**

**Jernej Završnik[1,4], jernej.zavrsnik@zd-mb.si**

**[1,4]Bojan Žlahtič [2], bojan.zlahtic@um.si**

**Grega Žlahtič[2], grega.zlahtic1@um.si**

**Institutions:**

1 *Community Healthcare Center Dr. Adolf Drolc Maribor, 2000 Maribor, Slovenia;*
2 *Faculty of Electrical Engineering and Computer Sciences, University of Maribor, 2000 Maribor, Slovenia*
3 *CMU- Africa, Bldg Plot No A8, 3594+2RG Kigali Special Economic, Regional ICT Center of Excellence, Kigali, Ruanda*
4 Alma Mater Europaea University, ECM Maribor, 2000 Maribor, Slovenia

* *Correspondence: Peter Kokol, peter.kokol@um.si*

**Background:** Climate change poses an existential threat to global public health, with Africa bearing a disproportionate burden of climate-sensitive health emergencies. While Artificial Intelligence (AI) and machine learning offer unprecedented capabilities for predictive modeling and early warning systems, their application within African public health frameworks remains under-documented and constrained. This study maps thematic divergences between global and African research landscapes to identify gaps in AI deployment for climate-health management.

**Methods:** Guided by a Synthetic Knowledge Synthesis approach, we conducted a scoping review and bibliometric analysis of Scopus-indexed literature (through October 2025). We utilized VOSviewer for bibliometric mapping, co-occurrence network analysis, and thematic clustering to structurally compare the Global Literature Corpus (GLC) and the African Literature Corpus (ALC).

**Results:** Network analysis reveals stark divergences in priorities. The GLC emphasizes complex, predictive planetary health applications, including air pollution mortality tracking and environmental modeling. Conversely, the ALC features localized, reactive solutions, primarily utilizing supervised machine learning for immediate infectious disease triage (e.g., HIV, malaria) and food security. The analysis identifies profound structural barriers in the ALC—notably a "compute deficit," fragmented data silos, and energy instability—precluding the scaling of advanced AI climate-health models.

**Conclusions:** A critical gap exists between global AI innovations and African public health realities. To build genuine resilience, international funding must pivot from siloed pilot projects toward investing in foundational digital infrastructure and localized, predictive AI frameworks tailored to the African context.

- **What is known about this subject:**
  - Global climate change is rapidly exacerbating public health crises, particularly in Africa where the burden of climate-sensitive infectious and non-communicable diseases is disproportionately high.
  - Artificial Intelligence and machine learning are highly effective tools for planetary health, capable of predicting disease outbreaks and optimizing resource allocation.
  - Advanced AI deployment in healthcare is currently concentrated in the Global North, trained on datasets that often exclude African populations and ecosystems.
- **What this study adds:**
  - It maps a deep thematic divergence: global research focuses on predictive environmental modeling, while African research is constrained to localized, reactive infectious disease triage.
  - It identifies specific structural barriers—a localized "compute deficit," data silos, and energy instability—preventing the deployment of advanced AI public health frameworks in Africa.
  - It outlines high-impact, contextually viable AI implementation strategies, such as AI-driven acoustic water leak detection for cholera prevention and predictive solar microgrid management

## BACKGROUND

The convergence of global climate change and health issues presents a rapidly escalating threat [1–6], demanding sophisticated predictive and adaptive strategies. This crisis is profoundly acute in Africa, where severe public health vulnerabilities are exposed and magnified by environmental shifts. The epidemiological reality is stark: Africa currently carries 95% of the global malaria burden, and climate-induced flooding directly spikes cholera outbreaks and rapidly expands vector habitats across the continent. Addressing these critical, compounding challenges requires moving healthcare systems from a reactive to a proactive approach, a transition where Artificial Intelligence (AI) is emerging as a transformative tool [7–11].

The health burdens discussed in this paper span two distinct disease categories that respond differently to climate change and require fundamentally different AI approaches. Communicable diseases, including vector-borne conditions such as malaria and dengue fever, and waterborne diseases such as cholera, are directly linked to shifts in temperature, rainfall, and ecosystem change that alter transmission pathways. Conversely, non-communicable diseases (NCDs), including cardiovascular disease and mental health conditions, are driven by indirect climate pathways such as worsening air quality, chronic heat stress, population displacement, and food insecurity. This distinction matters for AI application design: predictive models for vector-borne diseases rely on climate, mosquito habitat, and mobility data, while NCD-focused tools must integrate environmental exposure data (the "exposome") with population health records. Both categories impose a growing burden on African health systems. Recognising this distinction is essential for designing AI tools that are contextually appropriate; a satellite-based mosquito habitat model is of little use for addressing cardiovascular risk, and an exposome pipeline cannot predict a cholera outbreak.  **(see Figure 1)**.

AI's core utility lies in its capacity to process vast, high-dimensional datasets that conventional epidemiological models struggle to integrate [12]. Climate change is influencing the occurrence and distribution of diseases like malaria, dengue fever, and cholera by affecting vector habitats, water quality, and extreme weather events. AI can analyse complex variables, including temperature, precipitation patterns, humidity, land-use change, and human movement [13]. This integration allows AI models to outperform traditional statistical approaches by uncovering non-linear relationships and hidden patterns, and forecasting the spatial and temporal spread of outbreaks with higher accuracy [14].

Beyond the infectious disease spectrum, AI supports broader public health systems resilience by modelling the potential impacts of climate hazards like heatwaves and air pollution [15]. AI-driven geospatial analysis can identify specific vulnerable populations and hotspots disproportionately affected by environmental degradation, guiding the strategic allocation of scarce resources and targeted public health interventions [16]. Furthermore, AI and digital health contribute to climate mitigation in the energy sector—e.g., optimizing production and use renewable and clean energy—which indirectly improves public health by reducing reliance on fossil fuels and mitigating the associated air pollution risks [17–19].

According to above AI and digital health have emerged as transformational tools in planetary health. Globally, AI is being rapidly deployed to model complex environmental-health interactions, forecast outbreak trajectories, and optimize resource allocation during climate emergencies. However, the benefits of these digital health innovations are not equitably distributed. The development and deployment of advanced AI models are predominantly concentrated in the Global North, heavily trained on datasets that do not reflect African populations, ecosystems, or health systems.

Consequently, a "digital divide" threatens to widen the existing health disparities between high-income countries and the African continent. While global research pushes the boundaries of predictive environmental modeling and personal exposome tracking, the extent to which these technologies are being adapted for or researched within African contexts remains poorly documented. Furthermore, structural barriers—ranging from a lack of high-performance computing infrastructure (the "compute deficit") to data fragmentation and energy instability—complicate the translation of global AI innovations into localized African public health frameworks.

To address this critical knowledge gap, this study employs a bibliometric and scoping review approach to map the intersection of AI, climate change, and health. Specifically, we aim to systematically compare the Global Literature Corpus (GLC) with the African Literature Corpus (ALC) to identify thematic divergences, chart current research hotspots, and highlight structural barriers. By visualizing the landscape of current research, this review provides a strategic roadmap for researchers, policymakers, and international funding bodies to build equitable, AI-driven climate-health resilience in Africa.

## METHODOLOGY

Africa bears the heaviest, most disproportionate burden of the global climate crisis, manifesting through intensified phenomena like erratic precipitation, devastating floods, and chronic droughts [20]. Although the continent contributes negligibly to worldwide greenhouse gas output, its deep dependence on rain-fed agricultural systems and existing infrastructure deficits leave millions susceptible to climate related health issues or hazards [21]. Artificial Intelligence is rapidly proving to be a critical technological frontier, offering powerful, data-driven instruments for both climate adaptation and mitigation , and represents a significant, largely under-developed asset for building resilience across the African nations [18,22–24].

This study employed Synthetic Knowledge Synthesis (SKS) to discern differences in the research literature concerned with employment of artificial intelligence in the management of climate change-related health issues. SKS methodology, is a three-pronged analytical framework developed by Kokol to provide a comprehensive, semi-automated overview of the research landscape by analysing the whole corpora of publications and mostly biased and fragmented subsets [25]. SKS has already been successfully used in health care related research [26–31]. Because the SKS methodology analyzes the entire retrieved bibliographic corpus computationally, traditional PRISMA screening and eligibility exclusions of individual articles were not performed, eliminating manual selection bias. The PISMA diagram is shown in Figure 2.

The process begins with the extraction of primary bibliographic and meta data from the Scopus database, selected for its superior coverage of non-English literature, and robust metadata compared to alternative repositories. The SKS approach involves the triangulation of bibliometric mapping via VOSviewer to visualize the intellectual structure and thematic clusters of the field; and qualitative content analysis [32] to synthesize core findings and identify emerging research frontiers. By leveraging Scopus's extensive reach and SKS's scalable data processing, this methodology minimizes selection bias and efficiently manages "big data" corpora, facilitating an objective, reproducible, and holistic assessment of current scientific developments and future trajectories within the specified domain.

The analysis was executed on two distinct corpora:

- Global Literature Corpus (GLC): Publications concerned with the specified subject areas worldwide.
- African Literature Corpus (ALC): Publications specifically related to research within the African context.

Synthetic Knowledge Synthesis was applied independently in both corpora. Subsequently, a comparative analysis of the resulting SKS outputs was conducted. The identified disparities in research scope between the global and African literature were utilized to pinpoint potential research avenues or approaches that could be introduced and implemented effectively within the African environment and context.

The corpora were formed on 29$^{th}$ of April 2026 using the following two search strings, first for the GLC and the second for the ALC:

TITLE-ABS-KEY(( "artificial intelligence" OR "machine learning" OR "deep learning" OR "intelligent system" OR "support vector machine" OR ( "decision tree" AND ( induction OR heuristic ) ) OR "random forest" OR "Markov decision process" OR "hidden Markov model" OR "fuzzy logic" OR "k-nearest neighbour" OR "naive Bayes" OR "Bayesian learning" OR "artificial neural network" OR "convolution neural network" OR "recurrent neural network" OR "generative adversarial network" OR "deep belief network" OR "perceptron" OR {natural language processing} OR {natural language understanding} OR {general language model} ) and "climate change*" and health)

TITLE-ABS-KEY(( "artificial intelligence" OR "machine learning" OR "deep learning" OR "intelligent system" OR "support vector machine" OR ( "decision tree" AND ( induction OR heuristic ) ) OR "random forest" OR "Markov decision process" OR "hidden Markov model" OR "fuzzy logic" OR "k-nearest neighbour" OR "naive Bayes" OR "Bayesian learning" OR "artificial neural network" OR "convolution neural network" OR "recurrent neural network" OR "generative adversarial network" OR "deep belief network" OR "perceptron" OR {natural language processing} OR {natural language understanding} OR {general language model} ) and "climate change*" and health) AND TITLE-ABS-KEY(africa OR algeria OR angola OR benin OR botswana OR "Burkina Faso" OR burundi OR cameroon OR "Cape Verde" OR "Central African Republic" OR chad OR comoros OR congo OR "cote d*" OR "democratic republic off congo" OR djibouti OR egypt OR equator OR guinea OR eritrea OR swaziland OR ethiopia OR gabon OR gambia OR ghana OR guinea OR guinea-bissau OR kenya OR lesotho OR liberia OR libya OR madagascar OR malawi OR mali OR mauritania OR mauritius OR mayotte OR morocco OR mozambique OR namibia OR niger OR nigeria OR "réunion for rwanda" OR "Saint Helena" OR "São Tomé" OR "príncipe for senegal" OR seychelles OR "Sierra Leone" OR somalia OR "South Africa" OR "South Sudan" OR sudan OR tanzania OR togo OR tunisia OR uganda OR zambia OR zimbabwe)

Building upon the preliminary SKS employment, a subsequent phase of targeted desktop research and analysis was executed to identify high-impact Artificial Intelligence (AI) deployment opportunities within the African context that might have been omitted by the primary analysis. This secondary analysis specifically addresses the critical intersection of deteriorating infrastructure, pervasive energy poverty, and the persistent burden of both communicable and chronic diseases across Sub-Saharan Africa.

**RESULTS**

The resultant network (Figure 3), comprising associations (lines) between author keywords organized into discrete, colour-differentiated clusters, was subjected to further analysis. This analysis of inter-keyword links within clusters led to the identification of categories and the core research themes, as outlined in Table 1. To better align these technical themes with practical outcomes, a direct public health application metric has been applied to each cluster.

Repeating the approach above on the African Literature Corpus (ALC) publications resulted in the landscape shown in Figure 4. During the thematic clustering of the ALC, a minor cluster (Light blue, 10 keywords) focusing on the use of large language models to increase productivity in neurosurgery was identified. This cluster represents an unrelated overlap in the literature, resulting from the broad nature of the initial search string. Because it does not relate to climate change-induced public health impacts, it has been excluded from the subsequent thematic mapping in Table 2 to maintain the focus and integrity of the review.

The analysis of inter-keyword links within the remaining clusters led to the identification of categories, core research themes, and direct public health applications tailored to the African context, as listed in Table 2.

## DISCUSSION

### Statement of principal findings

This study highlights a profound thematic and operational divergence between global and African research landscapes regarding the application of Artificial Intelligence (AI) in climate-health management. Our network analysis of the Global Literature Corpus (GLC) reveals a strong emphasis on complex, predictive planetary health applications, such as air pollution mortality tracking, complex environmental modeling, and the integration of personal exposomes with chronic disease outcomes. In stark contrast, the African Literature Corpus (ALC) is characterized by localized, reactive solutions. The ALC predominantly utilizes supervised machine learning for immediate infectious disease triage (such as HIV and malaria) and basic sustainability measures. Ultimately, the deployment of advanced AI across Africa is fundamentally constrained by a localized "compute deficit," fragmented data silos, and energy instability. A more detailed discussion of possible application in Africa is presented bellow;

*Sea Temperature and Oceanography*

AI can provide crucial forecasting to protect coastal communities and vital 'Blue Economy' sectors.

- Early Warning Systems for Coastal Hazards: Use Machine Learning (ML) models trained on satellite data, buoy measurements, and historical data to predict extreme sea-level events, storm surges, and abnormal current patterns [33,34]. This directly informs disaster preparedness and coastal infrastructure protection in countries like South Africa and those along the East African coast.

- Fisheries Management: Apply computer vision and oceanographic data modeling to track fish migration patterns and marine ecosystem health (e.g., predicting coral bleaching events due to rising sea temperatures, as noted in Kenya's Blue Economy). This supports sustainable fishing quotas and protects the livelihoods of coastal communities [35,36].

- Climate Change Prediction: Leverage AI to improve the accuracy and high-resolution of local climate models by integrating data from various sparse African weather stations and global models, enabling better long-term adaptation strategies [37,38].

*Carbon Emission Prediction and Electric Vehicles (EVs)*

AI is key to leapfrogging to greener energy systems and mobility.

- Grid Optimization for Renewables: Use AI to forecast energy demand and optimize the integration of distributed renewable energy sources (solar, wind) into often-fragile African power grids, improving grid stability and reducing reliance on fossil fuels [39].

- Within African urban landscapes, Machine Learning facilitates the spatial optimization of EV charging infrastructure by analyzing population density, traffic dynamics, and grid resilience. This predictive modeling identifies strategic site locations, mitigating unique infrastructural bottlenecks to accelerate the continental transition toward sustainable, electrified mobility systems through evidence-based resource allocation. [40,41].

Aligning with Agenda 2063, AI-driven auditing enables real-time monitoring and predictive analytics for African industrial and transport emissions. By providing high-fidelity data, these tools empower policymakers to establish rigorous regulatory frameworks and enforce reduction targets. This technological integration is essential for harmonizing the continent’s industrial expansion with sustainable, climate-resilient development mandates [42].

*Health and Socio-Economic Impact*

While full-scale AI employment is challenged by infrastructure, AI can augment limited human resources.

- Teleradiology/Pathology: AI-powered computer vision tools can automatically analyse X-rays, blood smears (e.g., for malaria), and pathology slides with high accuracy. This is a form of robotization that reduces the workload on scarce specialists and makes expert diagnosis accessible in remote areas via digital health platforms [43,44].
- Surgical Training Simulators: Use augmented and virtual reality with AI feedback to provide low-cost, high-quality surgical training to medical professionals without needing complex physical labs [44]

Advanced AI can be used in Poverty and Malaria management. It offers predictive and prescriptive power for two of Africa's most persistent challenges.

- Poverty Targeting: Use satellite imagery (night-lights, roof materials, road networks) and mobile data to estimate poverty levels in granular detail. This advanced AI targeting ensures social welfare programs, food aid, and infrastructure development are directed precisely to the most vulnerable communities [45,46].
- Malaria, HIV, Tuberculosis, Chronic Obstructive Pulmonary Diseases and Cholera Outbreak Prediction: Integrate diverse data—climate (temperature, rainfall), mosquito breeding site detection via drones/images, and mobility data—into Deep Learning models to predict diseases outbreaks weeks in advance [9,47], enabling pre-emptive resource deployment (nets, medication, spraying) mitigating the immense socio-economic burden of outbreaks while enhancing the cost-effectiveness of public health expenditures across African sovereign budgets.

*Exposome and Specific Disease Associations (Cardiovascular, Mental)*

AI is necessary to handle the complexity of environmental health data.

- Personalized Risk Assessment: AI can integrate the exposome (all non-genetic environmental factors like air and water pollution, diet, stress) with local genomic data to identify specific environmental risk factors for non-communicable diseases [48] like Cardiovascular Disease (CVD) [48], which is rising in Africa.
- Mental Health Tele-triage: Employ Natural Language Processing (NLP) to analyse speech or text (where available in local languages) from digital health consultations to identify emotional distress and mental health risks, guiding users to appropriate (and often scarce) counselling resources [49].

*Digitalization and Disaster Response*

AI can transform disaster management from reactive to anticipatory.

- Flash Flood and Drought Forecasting: Utilize ML to rapidly process satellite, radar, and social media data to forecast localized flooding or accelerating drought conditions, providing early warnings tailored to specific communities [50,51].

- Damage Assessment: After a disaster (e.g., earthquake, cyclone), AI can rapidly analyse drone and satellite images to automatically map damaged infrastructure, prioritizing aid and rescue routes [52,53].

*Digital Health Innovation*

This theme is the foundational layer for AI adoption in healthcare.

- Interoperable Data Systems: Research into AI-driven tools to harmonize and link disparate health records (often paper-based or on separate digital systems) across clinics and regions, creating the unified, high-quality data sets needed to train local AI models effectively [54,55]

- Telemedicine Optimization: Use AI to intelligently match patients to the right virtual clinician or specialist, optimizing the limited time of medical experts [56,57].

However, implementing these foundational digital health innovations requires active engagement with the heterogeneity of infrastructure across the continent. Africa is not a monolith; the readiness to adopt and sustain AI-driven health tools varies enormously between countries. Rwanda, for instance, has a relatively mature digital health ecosystem, including drone-based medical supply delivery through Zipline, a national electronic medical records rollout coordinated through the Rwanda Biomedical Centre, and widespread DHIS2 adoption. Kenya has pioneered mHealth innovations and mobile-integrated health payment systems. Ethiopia operates one of the largest community health worker networks in the world, with structured digital data collection at scale. Ghana has similarly deployed drone delivery for vaccines and blood products. By contrast, many landlocked or conflict-affected countries face acute constraints in electricity access, mobile internet penetration, and local AI expertise that would fundamentally limit the applicability of the tools proposed here. Therefore, future iterations of this work and subsequent policy frameworks must consult country-level digital health readiness data to stratify recommendations by implementation context, rather than treating the continent as a single deployment environment.

*Sentiment Analysis in Extreme Health, Food Security, and Wildfires*

This provides real-time, ground-level intelligence on emerging crises.

- Food Insecurity Monitoring: Apply Sentiment Analysis and Topic Modeling to social media, local news articles, and citizen reporting (in local languages) to detect early conversations about rising food prices, scarcity, or distress, serving as a leading indicator of food insecurity before official data is available [58,59].
- Public Health Crisis Monitoring: Analyse online chatter to gauge public fear, trust in health authorities, or rumours during disease outbreaks, allowing health agencies to deploy targeted communication campaigns to counter misinformation and improve compliance [60,61].

*Additional opportunities identified by secondary analysis*

- 
- A primary lacuna identified in the initial analysis is the stabilization of water supply systems. In nations such as South Africa and Kenya, "non-revenue water"—water lost due to infrastructure decay, leaks, and unauthorized extraction—frequently reaches 40% [62]. AI-driven acoustic sensors and pressure-management algorithms offer a transformative solution by identifying the unique acoustic "signatures" of pipe bursts in real-time, allowing for rapid containment and repair [63]. Beyond economic recovery, such interventions are critical for disease prevention; stabilizing water pressure prevents the backflow of contaminated groundwater, a primary precursor to cholera epidemics. By integrating AI with Long Short-Term Memory (LSTM) networks that analyze rainfall anomalies and sanitation metrics, health agencies in Nigeria and Malawi can now forecast cholera outbreaks with over 80% accuracy, enabling the pre-emptive deployment of oral rehydration salts and vaccines [64]

- The "silent" infrastructure of energy security remains a fundamental barrier to health resilience. Approximately 600 million individuals in Sub-Saharan Africa lack grid access, compromising vaccine cold-chain integrity and hospital operations [65,66] AI is now being leveraged for geospatial site selection and load forecasting to optimize decentralized renewable energy (DRE) systems. In Kenya and Uganda, predictive algorithms manage "pay-as-you-go" solar microgrids, ensuring that rural clinics maintain continuous power for life-saving diagnostics while managing credit risk for unbanked populations [67,68].
- Furthermore, the management of high-burden diseases must be prioritized through localized AI training. While global models focus on chronic "exposome" data, African initiatives like *HealthPulse AI* and *Ubenytics* utilize computer vision on smartphones to interpret malaria blood smears with 98.5% accuracy, outperforming manual technician reviews in rural Africa [69]. Similarly, deep learning algorithms like *CAD4TB* are significantly reducing the diagnostic turnaround for Tuberculosis (TB) in South Africa and Ghana, while emerging pharmacometric AI models are being used to tailor dosing for HIV/TB co-infections to the specific genetic diversity of African populations . Finally, the rising prevalence of Chronic Obstructive Pulmonary Disease (COPD)—driven by household air pollution from biomass cooking fires—can be mitigated through AI-integrated air quality monitors that provide real-time health alerts to vulnerable communities [70].

**Strengths and weaknesses of the study**

A primary strength of this study is the deployment of the Synthetic Knowledge Synthesis (SKS) framework to evaluate the entire corpus of relevant literature in Scopus. By leveraging comprehensive knowledge synthesis rather than fragmented evaluations of different review types, this approach minimizes selection bias and effectively manages large bibliographic datasets. Furthermore, the dual-corpus methodology allows for an objective, reproducible visualization of the specific digital divides impacting public health. A weakness of this study is its reliance exclusively on English-language publications indexed in major databases, which may inadvertently omit localized, grassroots AI implementations documented in regional or non-English repositories.

**Strengths and weaknesses in relation to other studies**

While many existing studies outline the theoretical potential of AI in healthcare, this review directly interrogates the infrastructural realities of implementing these technologies in resource-constrained settings. Other bibliometric analyses frequently treat the global research landscape as a monolith. This study specifically isolates the African context, providing a highly granular assessment of systemic vulnerabilities such as unreliable early warning systems and energy poverty.

**Discussion of important differences in results**

The most critical difference identified in our results is the gap between predictive environmental capabilities in the Global North and the reactive application of AI in Africa. While global models focus heavily on chronic non-communicable diseases driven by indirect climate pathways, African applications are necessarily focused on immediate communicable disease burdens and basic food security. Furthermore, the analysis uncovers that the barrier to advanced AI adoption in Africa is not a lack of conceptual understanding, but rather structural deficits, including the lack of high-performance computing infrastructure and secure data harmonization systems.

## CONCLUSION

This study employed Synthetic Knowledge Synthesis to analyse the research landscape concerning AI use in climate change and health, successfully identifying a significant disparity in scope and thematic focus between global and African research scope. Globally, AI research is diverse, encompassing advanced applications in precision agriculture, air pollution modelling, sustainable development, and the 'exposome' of non-communicable diseases. In contrast, the African research focus, while vital, is currently narrower, concentrating on the use of supervised machine learning for specific diseases (e.g., HIV, malaria) and applications for basic sustainability and poverty reduction.

The comparative analysis reveals that advanced AI methodologies remain under-leveraged in Africa despite its disproportionate climate-health burden. This disparity arises from systemic infrastructural constraints—specifically energy instability, limited compute access, and data fragmentation—rather than a lack of expertise. Resolving these structural barriers is critical to de-risking AI investment and unlocking localized, tech-driven solutions that address the continent's unique environmental, infrastructural, and fiscal challenges. Addressing these barriers through targeted financial instruments, such as Green Bonds and Public-Private Partnerships, is critical to de-risking investment and unlocking localized AI-driven solutions that harmonize environmental resilience with fiscal sustainability. The identified prospective AI application themes—including AI-driven forecasting for coastal hazards, carbon emission prediction and EV infrastructure planning, teleradiology/pathology as a tool for human health resource augmentation, deep learning for poverty targeting and malaria outbreak prediction, and the use of sentiment analysis for food security monitoring—offer concrete pathways for accelerating climate change adaptation and strengthening public health resilience across Africa.

To effectively harness AI's transformative potential in Africa, future efforts must focus on capacity building and targeted investment. Specifically, there is an urgent need to:

1. Assure international and national funding research that adapts sophisticated global AI models (e.g., for exposome analysis, energy grid optimization, and disaster forecasting) to the unique and local data and infrastructure challenges prevalent in African nations.
2. International and multidisciplinary teams of AI researchers, climatologists, public health experts, and policymakers should ensure that technological advancements are contextually relevant and directly address the Africa's most pressing climate-health vulnerabilities.

In conclusion, the trajectory of AI adoption across the African continent is fundamentally constrained by deep-seated structural and resource-based inequities. The pervasive "compute deficit"—manifested in the scarcity of localized data centers and specialized hardware—not only exacerbates latency but undermines data sovereignty. Furthermore, chronic energy instability and the prohibitive costs of high-speed connectivity sustain a "usage gap" that stifles the scalability of indigenous innovation. Addressing these infrastructure bottlenecks is a prerequisite for economic agency; without strategic investment in sovereign compute and localized datasets, the continent risks a digital neocolonialism, remaining a consumer rather than a co-creator of global AI advancement.

DECLARATIONS

Ethics approval and consent to participate

Not applicable.

Consent for publication**

Not applicable.

Availability of data and materials**

The datasets analyzed during the current study are available from the corresponding author on reasonable request. Or state that they were generated using Scopus queries provided in Section 2.1).

Competing interests

The authors declare that they have no competing interests.

Funding

This research received no external funding.

Authors' contributions

PK, HBV, and BŽ conceptualized and supervised the study. PK, GŽ, and BŽ developed the methodology, analyzed the data, and performed the visualization. JZ and HBV reviewed and edited the manuscript. All authors read and approved the final manuscript.

Acknowledgements

Not applicable.

Legends

**Figure 1:** Conceptual framework illustrating the distinct pathways through which climate change drivers exacerbate both communicable and non-communicable disease categories, and their corresponding targeted Artificial Intelligence (AI) applications within public health systems.

**Figure 2:** PRISMA-adapted flow diagram detailing the Synthetic Knowledge Synthesis (SKS) methodology utilized in this study. The diagram illustrates the identification, screening, and inclusion phases for both the Global Literature Corpus (N=2,016) and the African Literature Corpus (N=274) retrieved from the Scopus database on April 29, 2026, comparing AI deployment in climate-health management.

**Figure 3:** VOSviewer bibliometric network map displaying the thematic clustering of author keywords within the Global Literature Corpus (GLC). The network is generated from 2,016 Scopus-indexed publications retrieved on April 29, 2026, representing the global research landscape concerning the application of artificial intelligence and machine learning in climate-health management.

**Figure 4:** VOSviewer bibliometric network map displaying the thematic clustering of author keywords within the African Literature Corpus (ALC). The network is generated from 274 Scopus-indexed publications retrieved on April 29, 2026, representing localized research applications of artificial intelligence in managing climate change-related health issues specifically within the African context.

**Table 1:** Thematic clusters, core research themes, and direct public health applications identified via Synthetic Knowledge Synthesis (SKS) within the Global Literature Corpus (GLC) of 2,016 Scopus-indexed publications (up to April 2026) regarding global artificial intelligence deployment in climate change-related health issues.

**Table 2:** Thematicsc clusters core research themes, and direct localized public health applications identified via Synthetic Knowledge Synthesis (SKS) within the African Literature Corpus (ALC) of 274 Scopus-indexed publications (up to April 2026) regarding artificial intelligence deployment for climate change-related health management in Africa.

Table 1

| Cluster colour and number of authors keywords shown in the research landscape | Theme induced using qualitative SKS analysis | Derived Direct Public Health Application/Impact |
|---|---|---|
| Blue cluster (23 author keywords) | Application of image processing and computer vision in smart/precision farming and agriculture supported by IoT | Mitigating malnutrition and food-insecurity-induced wasting/stunting in children under 5. |
| Dark blue cluster (32 author keywords) | Air pollution and morbidity and mortality | Reducing cardiovascular and respiratory disease mortality linked to climate-exacerbated air pollution and extreme weather events. |
| Red cluster (56 author keywords) | Remote sensing and google earth engine in managing climate change consequences on environment and health | Predicting and mitigating vector-borne disease outbreaks (e.g., malaria) and heat-related illnesses through early environmental surveillance. |
| Green cluster (34 author keywords) | Sustainable development during climate change for better health | Minimizing waterborne disease risks and improving urban respiratory health outcomes through resilient infrastructure and clean transport policies. |
| Yellow cluster (28 items) | Environment and digital public health | Preventing non-communicable diseases and managing climate-anxiety and mental health crises via targeted digital health triage. |
| Viollet cluster (27 author keywords) | Digitalization of health care and management of natural disasters | Enhancing emergency medical response and reducing trauma and mortality during climate-induced natural disasters. |

Table 2

| Cluster colour and number of authors keywords shown in the research landscape | Theme derived using qualitative SKS analysis | Derived Direct Public Health Application/Impact |
|---|---|---|
| Red cluster (25 keywords) | Machine learning application in increasing sustainability and reducing poverty | Targeting poverty-alleviation resources to highly vulnerable populations to reduce mal transmission and improve baseline community resilience. |
| Dark blue cluster (13 keywords) | Use of AI and satellite surveillance to in health innovation | Enhancing rural disease surveillance and directing mobile health interventions to remote mate-vulnerable communities. |
| Green cluster (22 keywords) | Application of machine learning for off-grid electrification, and for the prevention and control of diseases | Ensuring vaccine cold-chain integrity and powering rural primary healthcare clinics du extreme weather events. |
| Yellow cluster (14 keywords) | Application of deep learning for remote sensing and earth observation data analysis to increase food quality | Preventing climate-driven nutritional deficiencies and protecting rural food systems aga escalating drought conditions. |
| Viollet cluster (11 author keywords) | Analysis of earth and social media data for management of COVID-19 and ecosystems | Monitoring real-time public health crises and deploying rapid, targeted interventions ba on localized social intelligence. |
| Orange cluster (9 keywords) | Modelling of climate changes for increased food production | Securing staple crop yields to prevent localized famine, acute malnutrition, and associ opportunistic infections. |